\documentclass{article}
\usepackage{multirow}
\usepackage{longtable}
\usepackage{arxiv}
\usepackage{pdflscape}
\usepackage{subfig}
\usepackage{xcolor}
\usepackage{tabularx} 
\usepackage[utf8]{inputenc} 
\usepackage[T1]{fontenc}    
\usepackage[colorlinks=true, linkcolor=blue, citecolor=blue, urlcolor=blue, filecolor=blue, pdfborder={0 0 0}]{hyperref}
\usepackage{url}            
\usepackage{booktabs}       
\usepackage{amsfonts}       
\usepackage{nicefrac}       
\usepackage{microtype}      
\usepackage{lipsum}		
\usepackage{graphicx}
\usepackage{doi}
\usepackage{amsmath}
\usepackage{geometry}
\usepackage{array}
\usepackage{cleveref}
\usepackage{float}
\usepackage{lscape}
\usepackage{makecell}
\usepackage{adjustbox}
\usepackage{authblk}

 \title{RPEE-Heads: A Novel Benchmark for Pedestrian Head Detection in Crowd Videos}

 \author{Mohamad Abubaker\href{https://orcid.org/0009-0006-9119-4139}{\includegraphics[scale=0.06]{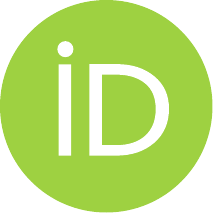}}$^{2}$, 
Zubayda Alsadder\href{https://orcid.org/0009-0008-2715-3345}{\includegraphics[scale=0.06]{orcid.pdf}}$^{2}$, 
Hamed Abdelhaq\href{https://orcid.org/0000-0003-4803-6689}{\includegraphics[scale=0.06]{orcid.pdf}}$^{2*}$,  
Maik Boltes\href{https://orcid.org/0000-0001-7240-896X}{\includegraphics[scale=0.06]{orcid.pdf}}$^{1*}$, \\
Ahmed Alia \href{https://orcid.org/0000-0002-3049-4924}{\includegraphics[scale=0.06]{orcid.pdf}}$^{1,2*}$}

\affil{
  $^{1}$ Institute for Advanced Simulation, Forschungszentrum Jülich, 52425 Jülich, Germany \\
 {m.boltes@fz-juelich.de}, {a.alia@fz-juelich.de}
}
\affil{
  $^{2}$ Department of Information Technology, An-Najah National University, P4110257 Nablus, Palestine \\
 {hamed@najah.edu}
}

\renewcommand{\shorttitle}{\textit{} }

\hypersetup{
pdftitle={A template for the arxiv style},
pdfsubject={q-bio.NC, q-bio.QM},
pdfauthor={David S.~Hippocampus, Elias D.~Striatum},
pdfkeywords={Deep Learning, Computer Vision, Object Detection},
}

\begin{document}
\maketitle

\begin{abstract}
The automatic detection of pedestrian heads in crowded environments is essential for crowd analysis and management tasks, particularly in high-risk settings such as railway platforms and event entrances. These environments, characterized by dense crowds and dynamic movements, are underrepresented in public datasets, posing challenges for existing deep learning models. 
To address this gap, we introduce the Railway Platforms and Event Entrances-Heads (RPEE-Heads) dataset, a novel, diverse, high-resolution, and accurately annotated resource. It includes 109,913 annotated pedestrian heads across 1,886 images from 66 video recordings, with an average of 56.2 heads per image. Annotations include bounding boxes for visible head regions. 
In addition to introducing the RPEE-Heads dataset, this paper evaluates eight state-of-the-art object detection algorithms using the RPEE-Heads dataset and analyzes the impact of head size on detection accuracy. The experimental results show that You Only Look Once v9 and Real-Time Detection Transformer outperform the other algorithms, achieving mean average precisions of 90.7\% and 90.8\%, with inference times of 11 and 14 milliseconds, respectively. Moreover, the findings underscore the need for specialized datasets like RPEE-Heads for training and evaluating accurate models for head detection in railway platforms and event entrances. The dataset and pretrained models are available at \url{https://doi.org/10.34735/ped.2024.2}.

\end{abstract}

\section{Introduction}
 
Detecting pedestrians in videos of crowded environments has tremendous significance for effectively understanding and managing such crowds, with several real-world applications, including pedestrian tracking~\cite{brunetti2018computer}, crowd counting~\cite{deng2023deep}, trajectory extraction~\cite{lu2017trajectory}, density estimation~\cite{fan2022survey}, and abnormal behavior detection~\cite{alia2024novel, khekan2024impact, alia2023cloud, alia2022hybrid}.
With rapid urbanization, dense crowds have become widespread in various locations~\cite{alia2022exploitation}, such as event entrances and railway platforms, where it is encountered daily, often leading to comfort and safety risks~\cite{ocejo2014subway, alia2022fast}.    However, detecting individual pedestrians becomes significantly more complex as crowd density increases due to frequent partial or complete occlusions. To alleviate this problem, researchers have started focusing on localizing the most visible part of the human body in such crowds—the head~\cite{zhou2024unihead, lu2020semantic}. Nevertheless, detecting person heads remains challenging due to variability in sizes, poses, and appearances of heads, as well as cluttered and dynamic backgrounds, varying lighting conditions, and occlusions~\cite{khan2021scale}.

Automatic head detection falls within the realm of computer vision, specifically in the object detection domain. With the rapid development of Deep Learning (DL), algorithms based on Convolutional Neural Networks (CNN)~\cite{dhillon2020convolutional} have achieved remarkable success in this domain. 
One of the critical reasons for this success is that CNN can automatically learn relevant features~\cite{alia2021enhanced} from data without human intervention~\cite{singh2021convolutional}. You Only Look Once (YOLO)~\cite{wang2023yolov7, yolov8, wang2024yolov9}, Region (R)-CNN, Fast R-CNN, Faster R-CNN, and Cascade R-CNN~\cite{girshick2014rich, girshick2015fast, ren2015faster, cai2018cascade} are popular CNN-based algorithms. While deep CNN-based algorithms are powerful, the performance of object detection models also relies heavily on the availability of large, diverse datasets with precise annotations, including bounding boxes around the target objects. However, although some publicly available datasets are available for head detection, there is an extreme scarcity of datasets specifically suitable for head detection in crowds within railway platforms and event entrances. For instance, the SCUT-HEAD (South China University of Technology Head)~\cite{peng2018detecting} dataset, derived from images of students in classrooms, and the Hollywood dataset~\cite{vu15heads}, sourced from movie scenes, differ significantly from real-world crowd scenarios. Additionally, the NWPU-Crowd (Northwestern Polytechnical University Crowd)~\cite{Wang2021} and JHU-CROWD++ (Johns Hopkins University Crowd++)~\cite{Sindagi2022} datasets contain images with very small heads, lacking the detail needed for advanced DL models to learn effectively~\cite{peng2024maritime}. Furthermore, datasets like Mall~\cite{chen2012feature}, SmartCity~\cite{zhang2018crowd}, and Train Station~\cite{farhood2017counting} provide only point-level annotations for heads in crowds, rather than the bounding boxes required for accurate head detection.
The lack of datasets with head bounding box annotations that cover the diverse and complex dense crowd scenarios at railway platforms and event entrances hinders the development of robust and accurate head detection algorithms in these environments.
 
To address the above limitation, we introduce the RPEE-Heads dataset, specifically designed for head detection in crowded environments, focusing on scenarios at Railway Platforms and Event Entrances.
This dataset is created to enhance the robustness and generalization capabilities of head detection models by providing a diverse and comprehensive collection of 1,886 annotated images, featuring 109,913 bounding boxes. On average, each image contains 56.2 head annotations. Moreover, the dataset includes a diverse range of scenes, covering indoor and outdoor environments, different seasons, weather conditions, varying levels of illumination, head scales, and appearances. It additionally features various resolutions and crowd densities, captured during day and night from multiple viewing angles—front, top, side, and back. 
Furthermore, the RPEE-Heads dataset includes detailed annotations with bounding boxes for the visible regions of heads, significantly contributing to the training and evaluation of advanced head detection algorithms.

The contribution of this paper can be summarized as follows.
\begin{enumerate}
    \item This paper introduces the first image dataset specifically focused on head detection for railway platforms and event entrance scenarios. This dataset facilitates the development of accurate machine learning and DL models for head detection in these critical environments, which are essential for a wide range of crowd safety applications.
     
    \item It performs a thorough empirical analysis, comparing eight state-of-the-art DL detection algorithms across multiple publicly available image datasets --annotated with head bounding boxes-- in addition to the newly introduced dataset. This analysis and the RPEE-Heads dataset provide the research community with a solid baseline for further advancements and improvements in head detection. 
    
    \item This paper presents an empirical study on head size's impact on detection algorithms' accuracy.
\end{enumerate}

The rest of this paper is structured as follows. In the beginning, we review the literature to explore different benchmark contributions. Afterward,~\cref{sec:dataset} details RPEE-Heads in terms of data sources, annotation process, and dataset creation. Then,   experimental results and comparisons are discussed in~\cref{sec:experiments}. Finally, Section~\ref{sec:conclusion} concludes the paper.

\section{Related Work}
In this section, we first provide a brief overview of popular deep CNN-based object detection algorithms, followed by a review of DL models for pedestrian head detection. Finally, we discuss the most relevant public datasets annotated with pedestrian head bounding boxes, along with their limitations.  

\subsection{DL-based Object Detection Algorithms}

DL, especially CNN-based algorithms, has recently advanced object detection in images and videos.  Most of such algorithms can be divided into two-stage~\cite{girshick2014rich, girshick2015fast, ren2015faster, cai2018cascade}] and single-stage~\cite{wang2023yolov7, wang2024yolov9, yolov8, lin2017focal} categories. Two-stage algorithms, such as R-CNN~\cite{girshick2014rich}, initially extract candidate object regions and then classify them into specific object classes. To reduce computational time in R-CNN, Fast R-CNN~\cite{girshick2015fast} optimizes feature extraction by computing features for the entire image at once rather than separately for each candidate region. Building on Fast R-CNN, Faster R-CNN~\cite{ren2015faster} introduced Region Proposal Networks to speed up detection further. To address the scale-variance issue in such algorithms, Cascade R-CNN~\cite{cai2018cascade} employed a series of detectors with progressively increasing Intersection over Union thresholds. 

On the other hand, single-stage algorithms enhance detection speed by predicting bounding boxes and class probabilities in a single pass. Recent advancements in this area include You Only Look Once (YOLO) v7x~\cite{wang2023yolov7}, YOLOv8x~\cite{yolov8}, and YOLOv9-E~\cite{wang2024yolov9}, which have achieved significant improvements in both speed and accuracy. Each version of YOLO has been introduced to enhance both  speed and accuracy in object detection. Another example of single-stage algorithms is RetinaNet-101~\cite{lin2017focal}, which addresses the challenge of class imbalance during training.

With the advancements in object detection driven by efficient real-time YOLO versions, a new contender, Real-Time Detection Transformer (RT-DETR)~\cite{zhao2023detrs}, has emerged, leveraging vision transformer technology to push the field forward.

\subsection{DL-based Head Detection Models}

With the tremendous success of CNNs for object detection, most of the current efficient methods for detecting pedestrian heads in images and videos are based on CNNs. For instance, Vu et al.~\cite{vu2015context}  and Li et al.~\cite{li2016localized} developed two models based on R-CNN. Similarly, a custom CNN model was introduced in Ref.~\cite{chen2018headnet}. Another approach, proposed by Wang et al.~\cite{wang2017robust}, proposed a model based on the Single Shot MultiBox Detector.  In the same direction, Khan et al.~\cite{khan2019disam}  presented a new model combining CNNs with scale-aware head proposals.
Another example, YOLOv5 with a transfer learning technique, was employed in a new model in Ref.~\cite{hassan2023crowd}. Additionally, Vo et al.~\cite{vo2022pedestrian} developed a new head detection model based on encoder-decoder Transformer networks.
Continuing this trend of integrating advanced architectures, study~\cite{zhou2024unihead} proposed an approach combining deep convolution, Transformer, and attention mechanisms.

The above models were often tailored to specific scenarios because the training datasets do not adequately cover the diverse and complex conditions across all crowded environments, such as event entrances and railway platforms. The following section will review several popular datasets with head annotations and highlight their limitations in the contexts of event entrances and railway platforms.

\subsection{Datasets Annotated with Pedestrian Head Bounding Boxes}

Several image/video-based datasets with pedestrian head bounding box annotations have been introduced in the literature to advance applications in crowd dynamics. This section reviews various such datasets, including NWPU-Crowd~\cite{Wang2021}, JHU-CROWD++~\cite{Sindagi2022}, Hollywood Heads~\cite{vu15heads}, SCUT-Head Part B~\cite{peng2018detecting}, SCUT-Head Part A~\cite{peng2018detecting}, FDST (Fudan-ShanghaiTech)~\cite{Fang2019}, CroHD (Crowd High-Definition)~\cite{sundararaman2021tracking}, and CrowdHuman~\cite{shao2018crowdhuman}. Table~\ref{tab:datasets} provides a summary of these datasets.

\begin{table}[ht]
\centering
\caption{Characteristics of the related datasets.}
\label{tab:datasets_related_work_part1}
\footnotesize
  
\begin{tabular}{l c c c c c c c }
\toprule
\textbf{Dataset} & \textbf{Year}
& \textbf{\makecell {Frames \\ Count}} & \textbf{ \makecell{Average Resolution \\ W $\times$ H (pixels)} }& \textbf{\makecell{Total \\ Heads}} &
 \textbf{\makecell{Average \\ Heads}} & \textbf{\makecell{Max \\ Heads}} & 
\textbf{Scene Description}    \\ \midrule
FDST ~\cite{Fang2019} & 2019
& 15,000 & \makecell{1,920 $\times$ 1,080, \\ 1,280 $\times$ 720} & 394,081 &  26.3 & 57 & \makecell{Inside Malls, \\ Streets and Public Spaces}  \\
SCUT-HEAD Part A ~\cite{peng2018detecting} &2018
& 2,000 & 1,076 $\times$ 605 *& 67,321 & 33.6 & 88 &  Classroom    \\
SCUT-HEAD Part B ~\cite{peng2018detecting} & 2018
& 2,405 & 994 $\times$ 675 *& 43,930 &  18.26 & 180 &  Classroom  \\
JHU-CROWD++ ~\cite{Sindagi2022} & 2022
& 4,372 & 1,430 $\times$ 910 & 1,515,005 &  346.5 & 25,791 & \makecell{Stadiums, Streets,\\ Concerts, Protests, etc.}    \\
CrowdHuman~\cite{shao2018crowdhuman} &2018
& 24,370 & 1,361 $\times$ 967 *& 339,565 &  22.64 & 391 & \makecell{Roads, Parties\\ Playing Basketball, \\  and Transportation Modes}   \\ 
CroHD~\cite{sundararaman2021tracking} & 2021
& 11,463 & 1,920 $\times$ 1,080 & 2,276,838 &  178 & 346 & Train Station and Streets   \\
NWPU-crowd ~\cite{Wang2021} & 2021
& 5,109 & 3,383 $\times$ 2,311 & 2,133,238 &  418.5 & 20,033 & \makecell{Stadium, Conference,\\ Street, Campus, Mall,\\ Museum, Station}  \\
Hollywood Heads ~\cite{vu15heads} & 2015
& 224,740 & 590 $\times$ 326 *& 369,846 &  1.6 & 27 & Hollywood Movies \\
\bottomrule
\multicolumn{8}{p{400pt}}{\makecell[l]{
$^*$  The resolution is determined as the average across all images in the dataset due to the varying resolutions between them.}}
\end{tabular}
\label{tab:datasets}
\end{table}

The JHU-CROWD++ and NWPU-Crowd datasets are characterized by highly dense crowds, containing up to 25,791 and 20,033 head annotations per image, respectively. JHU-CROWD++ offers approximately 1.5 million head annotations at a resolution of 1430 $\times$ 910, while NWPU-Crowd includes over 2 million head annotations with a high image resolution of 3,383 $\times$ 2,311. the CroHD dataset also comprises 11,463 frames of densely packed crowds, averaging 178 heads per image at a resolution of 1,920 $\times$ 1,080, totaling over 2 million annotated heads. In contrast, the CrowdHuman dataset features relatively lower crowd densities, with head counts ranging from 1 to 391 per frame, and contains approximately 339,566 head annotations.
Although these datasets are large, richly annotated, and encompass diverse scenarios with dense crowds, they could not be efficient for training detection algorithms. A significant issue is the prevalence of small, visible heads against cluttered dynamic backgrounds within these datasets. Typically, a head is considered small if its size adversely affects the feature extraction process, leading to suboptimal performance in detection algorithms~\cite{zhang2016far}. Section~\ref{sec:headsize} will explore the impact of head size on the performance of advanced detection algorithms.

Conversely, the Hollywood Heads, SCUT-Head Part B, SCUT-Head Part A, and FDST datasets contain a few small heads, which can aid in training advanced object detection algorithms such as DL models.
However, more than head size is needed to guarantee the development of accurate models for head detection in scenarios like event entrances and railway platforms; dataset diversity, large size, and the presence of similar or near scenarios are also crucial factors.
For instance, the Hollywood Heads dataset includes annotations for 369,846 human heads across 224,740 movie frames from Hollywood films, with an average of 1.6 heads per image.
Unfortunately, most frames in this dataset feature only one or two individuals, rather than the crowds typically found at event entrances or railway platforms.
SCUT-HEAD is another example, comprising 4,405 images labeled with 111,251 head annotations. The dataset has two parts: Part A includes 2,000 images sampled from classroom monitor videos at a university, containing 67,321 annotated heads with an average of 33.6 heads per image. Part B consists of 2,405 images captured from the Internet, with 43,930 annotated heads, averaging 18.26 heads per image.
However, such a dataset lacks diversity regarding indoor/outdoor environments, scenarios, the weather conditions, occlusions, and lighting variations.
To enhance diversity, Fang et al. introduced the FDST dataset collected from 13 different scenes, including shopping malls, squares, and hospitals. It comprises 15,000 frames with 394,081 annotated heads. 
Yet, despite the variety of scenes, the dataset may still lack sufficient diversity in weather and lighting conditions, as well as instances of complex occlusions and interactions between individuals. 

In summary, the discussed datasets may not be efficient for training and evaluating detection algorithms to accurately identify pedestrian heads in crowded environments, such as event entrances and railway platforms, due to the following limitations: 1) Some datasets include many small heads, often with limited relevant features. 2) Some datasets include scenarios not representative of the dense crowds at event entrances and railway platforms. 3) All datasets lack diversity in some critical aspects, such as camera angles, weather conditions, indoor and outdoor environments, day and nighttime, seasons, lighting conditions, head scales, crowd levels, and resolutions.
This paper introduces a novel dataset with head bounding box annotations to address these limitations. It additionally conducts an empirical comparative study of several state-of-the-art DL algorithms on the new dataset and existing public datasets. 
The following section introduces the dataset.

\section{RPEE-Heads Dataset}
\label{sec:dataset}
This section aims to describe the diverse high-resolution RPEE-Heads dataset. The details of the data sources, data annotation, and dataset creation are provided below.

\subsection{Data Sources}
\begin{figure}[H] 
\centering
\includegraphics[width=0.8\linewidth]{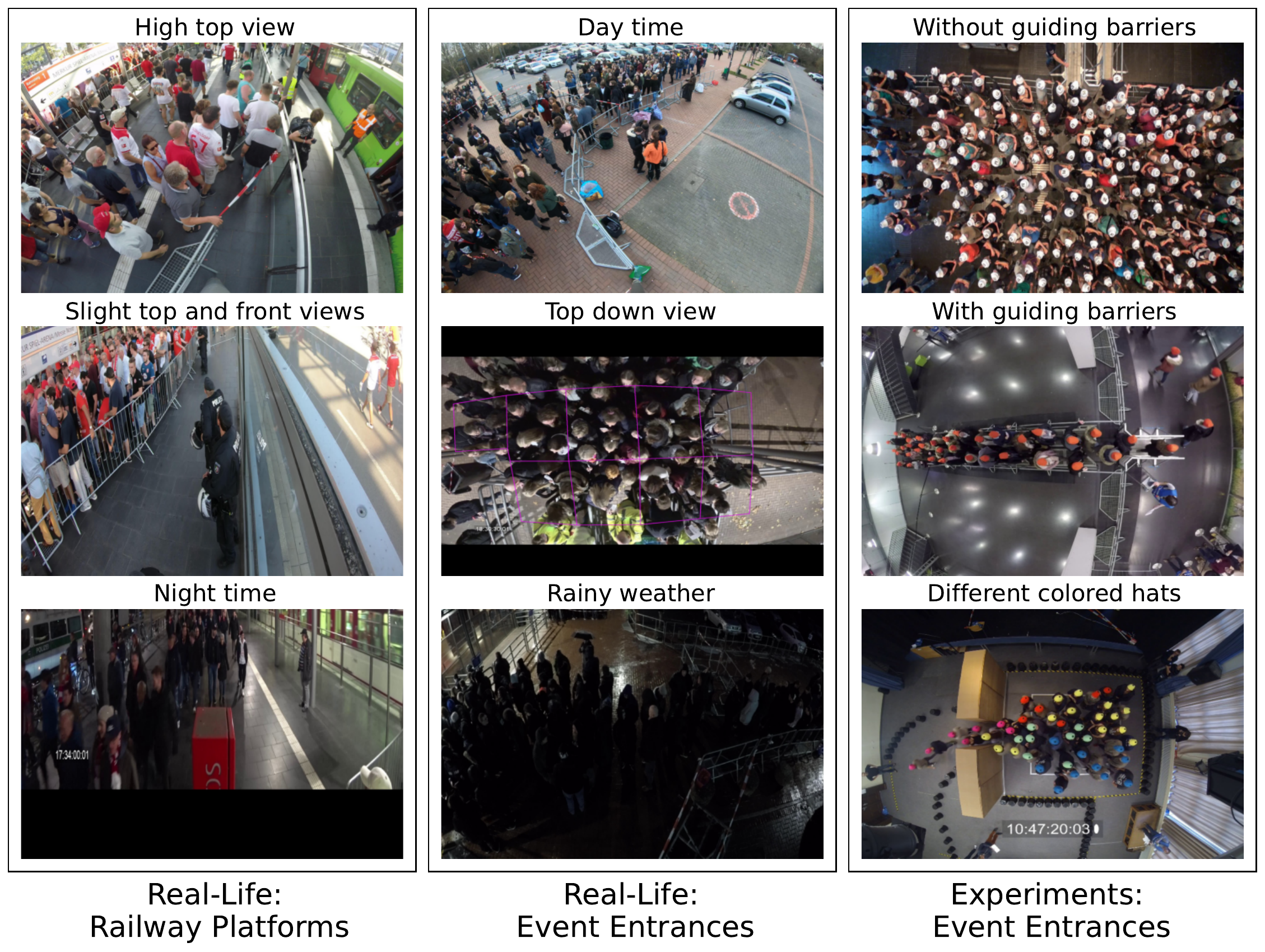}
    \caption{Illustrative examples from the selected data sources.}
    \label{fig:datasource}
\end{figure}

A total of 66 video recordings were selected to enrich the diversity of the proposed dataset, which focuses on railway platforms and event entrances. As shown in~\cref{tab:datasources} and~\cref{fig:datasource}, it incorporates a wide range of real-life scenarios and experiments, offering diversity in viewpoints, lighting conditions, weather conditions, indoor and outdoor environments, head sizes, and frame resolutions. The sources include:

\begin{enumerate}
  \item \textbf{Railway platforms:} 15 videos, recorded for research and educational purposes as part of the CroMa project~\cite{CroMA}, were selected. They were gathered from Merkur Spiel-Arena/Messe Nord station in Düsseldorf across various months and timeframes. The footage encompasses high-top, slightly-top, side, back, and front views, utilizing cameras with a frame rate of 25 frames per second and diverse resolutions. Additionally, the scenes contain pedestrians ranging from 6 to 132. Table \ref{tab:datasources} illustrates more details about such a data source, and the first column of~\cref{fig:datasource} includes examples from railway platforms.
  
  \item \textbf{Music concert entrances:}  We selected 34 video recordings, which were collected using multi-viewpoint cameras at a frame rate of 25 frames per second, and various resolutions for educational and research purposes within the CroMa Project~\cite{CroMA}. These recordings, including data from daytime and nighttime, averaged about 58 individuals per scene. Additional details can be found in~\cref{tab:datasources} and~\cref{fig:datasource}.

  \item \textbf{Real-world event entrance experiments:} We have selected 17 video experiments from the open data archive at Forschungszentrum Jülich, available under a CC Attribution 4.0 International license~\cite{crowdqueue, sieben2017collective,   adrian2020crowds, adrian2020influence, garcimartin2018redefining}. These experiments simulate crowded event entrances, with 8 involving guiding barriers, and 9 without. Such videos were recorded by a top-view static camera at a resolution of 1,920 $\times$ 1,440 or 1,920 $\times$ 1,080 pixels and a frame rate of 25 frames per second.  Examples of overhead views of some experiments are depicted in~\cref{fig:datasource}, while~\cref{tab:datasources} provides detailed characteristics of the selected experiments.
\end{enumerate}

\subsection{Data Annotation and Dataset Creation }

Creating the annotated pedestrian head dataset (RPEE-Heads) at railway platforms and event entrances involves three steps. The first step, frame selection, aims to extract the frame sequence $F$ from each video $V$ based on the corresponding frame interval $\Delta f$, and $F$ is described as follows:

\begin{equation}
  F=\{f_t \, | \,  t=1,\Delta f , 2\Delta f, 3\Delta f,4\Delta f, \dots,   T \}  ,  
\end{equation}

where $t$ is the order of the frame $f$ in $V$, and $T$ is the total number of frames in $V$.
The primary objective of selecting an image at each $\Delta f$ in the associated $V$ is to minimize the occurrence of duplicate scenes while preserving valuable information, thereby enhancing the dataset's quality. Table~\ref{tab:datasources} displays the values of $\Delta f$ for each video, along with the corresponding number of extracted frames. 
More specifically, a total of 1,886 frames were extracted from a duration of 32,369 seconds.

\begin{landscape}
\begin{table} 
\caption{Characteristics of the data sources.}
\footnotesize	
\centering
\begin{tabular}{lcccccccccccc}
\toprule
  Scenario &
 Videos &
  Duration (s) &
  \makecell[c]{Viewpoint\\ Vertical, Horizontal$^*$     }&
 
  \makecell[c]{Extracted \\ Frames} &
   \makecell[c]{Frame \\ Interval } &
\makecell[c]{Frame \\ Resolution} &
  \makecell[c]{Min \\ Heads} &
  \makecell[c]{Max \\ Heads} &
  \makecell[c]{Average \\ of Heads} &
  \makecell[c]{Total \\ Heads} &
  \makecell[c]{Daytime,  \\ Nighttime} &
  \makecell[c]{Indoor,  \\ Outdoor} \\ \midrule

\multirow{5}{*}{\begin{tabular}[c]{@{}l@{}}Railway \\ Platforms$^{**}$  \end{tabular}}  &
  5 &
  3,545 &
  High top, back &
  115 &
  30 &
  2,704 $\times$ 2,028 &
  26 &
  95 &
  46.6&
  5,368 &
  Day &
  Outdoor \\  
 
   &
  4 &
  2,124 &
  Slightly top, front &
  64 &
  30 &
  4,000 $\times$ 3,000 &
  6 &
  67 &
  53.9 &
  3,455 &
  Day &
  Outdoor \\ 
  & 
   1 &
  709 &
  High top, front &
  35 &
  20 &
  2,704 $\times$ 2,028 &
  22 &
  47 &
  52.2 &
  1,826 &
  Day &
  Outdoor \\ 
 &

  2  &
  1,062 &
  High top, side &
  52 &
  20 &
  3,840 $\times$ 2,160 &
  80 &
  132 &
  102.9 &
  5,345 &
  Day &
  Outdoor \\ 
 &
  
  3 &
  2,160 &
  Slightly top, front &
  48 &
  60 &
  4,000 $\times$ 3,000 &
  9 &
  101 &
  40.3 &
  1,933 &
  Night &
  Outdoor \\ 
  \\
   Avg/Total& 

  15  &
  9,600 &
  Multiple &
  314 &
  32 &
  Multiple &
  28.6 &
  88.4 &
  57.1 &
  17,927 &
  
  Multiple &
  Outdoor
  \\ \midrule
\multirow{5}{*}{\begin{tabular}[c]{@{}l@{}} Music Concert \\ Entrances $^{***}$\end{tabular}} &

  15 &
  8,512 &
   High top, front &
  452 &
  30 &
  4,000 $\times$ 3,000 &
  17 &
  201 &
  60.5 &
  26,093 &
 
  Both &
  Outdoor \\ 
  &
  3 &
  1,950 &
  High top, side &
  63 &
  30 &
  2,704 $\times$ 2,028 &
  23 &
  152 &
  56.1 &
  4,911 &
  Day &
  Outdoor \\ 
  &
 10  &
  5,320 &
   High top, side &
  346 &
  20 &
  3,840 $\times$ 2,160 &
  12 &
  91 &
  53.5 &
  18,508 &
  Both &
  Outdoor \\ 
  &
  6 &
  5,130 &
  Top down &
  132 &
  30 or 60 &
  4,000 $\times$ 3,000 &
  1 &
  136 &
  59.9 &
  7,974 &
  Both &
  Outdoor \\ 
  
  \\ 
  Avg/Total &
  34 &
  20,912 &
  Multiple &
  993 &
  34 &
  Multiple &
  13 &
  129.7 &
  57.8 &
  57,486 &
  Multiple &

  Outdoor  \\ \midrule
\multirow{2}{*}{\begin{tabular}[c]{@{}l@{}}Event Entrances \\ Experiments \end{tabular}} &
   8$^{****}$  &
  1,094 &
  Top down &
  396 &
  3 &
  1,920 $\times$ 1,080 &
  3 &
  172 &
  52.3 &
  21,125 &

  --- &
  Indoor \\ 
 &
  9$^{*****}$ &
  763 &
  Top down &
  183 &
  3 &
  1,920 $\times$ 1,440 &
  4 &
  255 &
  79.2 &
  13,375 &
  
  --- &
  Indoor \\  
  \\ 
   Avg/Total &
  
  17 &
  1,857 &
  Top down &
  579 &
  3 &
  Multiple &
  3.5 &
  213.5 &
  58.5 &
  34,500 &
  
  --- &
  Indoor \\
  
  \\  \cline{1-13} \\
   Overall (Avg/Total)&
   66 &
    32,369 &
    Multiple &
    1,886 &
   28 &
    Multiple &
    16.9 &
    121.9 &
    56.2 &
   109,913 &
  
    Multiple &
    Multiple
    \\ \\
  \bottomrule
  
  \multicolumn{13}{p{600pt}}{\makecell[l]{
  Min Heads refers to the minimum number of heads in a single frame. \\
  Max Heads represents the maximum number of heads present in a single frame. \\
  Average of Heads is  the average number of heads per frame. \\
  Total Heads indicates the total number of heads across all frames. \\
  $^*$ The horizontal position that encompasses the majority of the human crowd.\\
  $^{**}$ The data was collected at Merkur Spiel-Arena/Messe Nord train station in Düsseldorf, Germany on various dates: August 24, 2019; October 19, 2019; and November 3, 2019. \\ 
  $^{***}$ The data was recorded at four different concerts held at Mitsubishi Hall in Düsseldorf, Germany, on the following dates: March 19, 2019; July 11, 2019; January 25, 2019; and \\ November 29, 2019. \\
   $^{****}$ The entrance setups include guiding barriers.\\
    $^{*****}$ The setups mimic entrances that have guiding barriers.}
  }\\ 
\end{tabular}
\label{tab:datasources}
\end{table}
\end{landscape}

\begin{figure}[!ht]
    \centering
\includegraphics[width=0.8\linewidth]{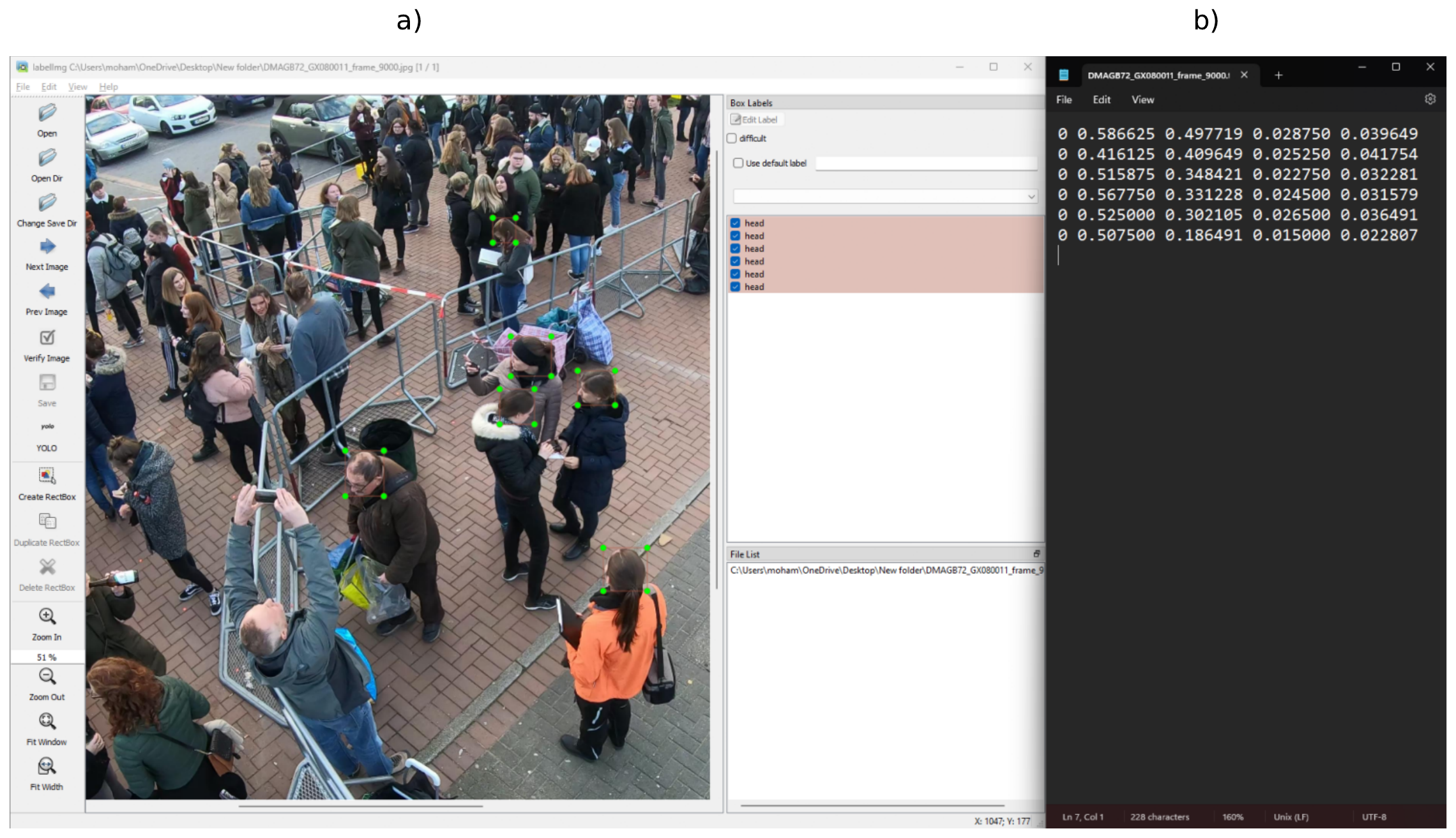}
    \caption{a) A visual example of manual annotation, with red boxes and green corners indicating head annotations.  b) The corresponding annotation text file. }
    \label{fig:annotataionexample}
\end{figure} 

In the second step, accurate annotations were manually added to the selected frames using the LabelImg tool, an open-source graphical image annotation tool~\cite{Tzutalin2015}. Each annotation represents a bounding box around the visible part of the head, and the information of all annotations for each image was stored in a corresponding text file. Figure~\ref{fig:annotataionexample} provides an illustrative example of an annotated frame alongside the related labeled file. Such a file is formatted with one row for each head (box) $\textless class, x, y, w, h\textgreater$, where $class$ is the label of the annotated object. The coordinates $x$ and $y$ specify the center of the box, while $w$ and $h$ indicate the width and height of the box, respectively. The values of $x$, $y$, $w$, and $h$ are given in pixel units and normalized from 0 to 1.
For normalization, the following equations are used:

\begin{equation}
    normalized\_x=\frac{x}{frame\_width}
\end{equation}

\begin{equation}
    normalized\_y=\frac{y}{frame\_height}
\end{equation}

\begin{equation}
     normalized\_w=\frac{width}{frame\_width}
\end{equation}

\begin{equation}
    normalized\_h=\frac{height}{frame\_height}
\end{equation}

\begin{figure}[h!]
    \centering \includegraphics[width=0.6\linewidth]{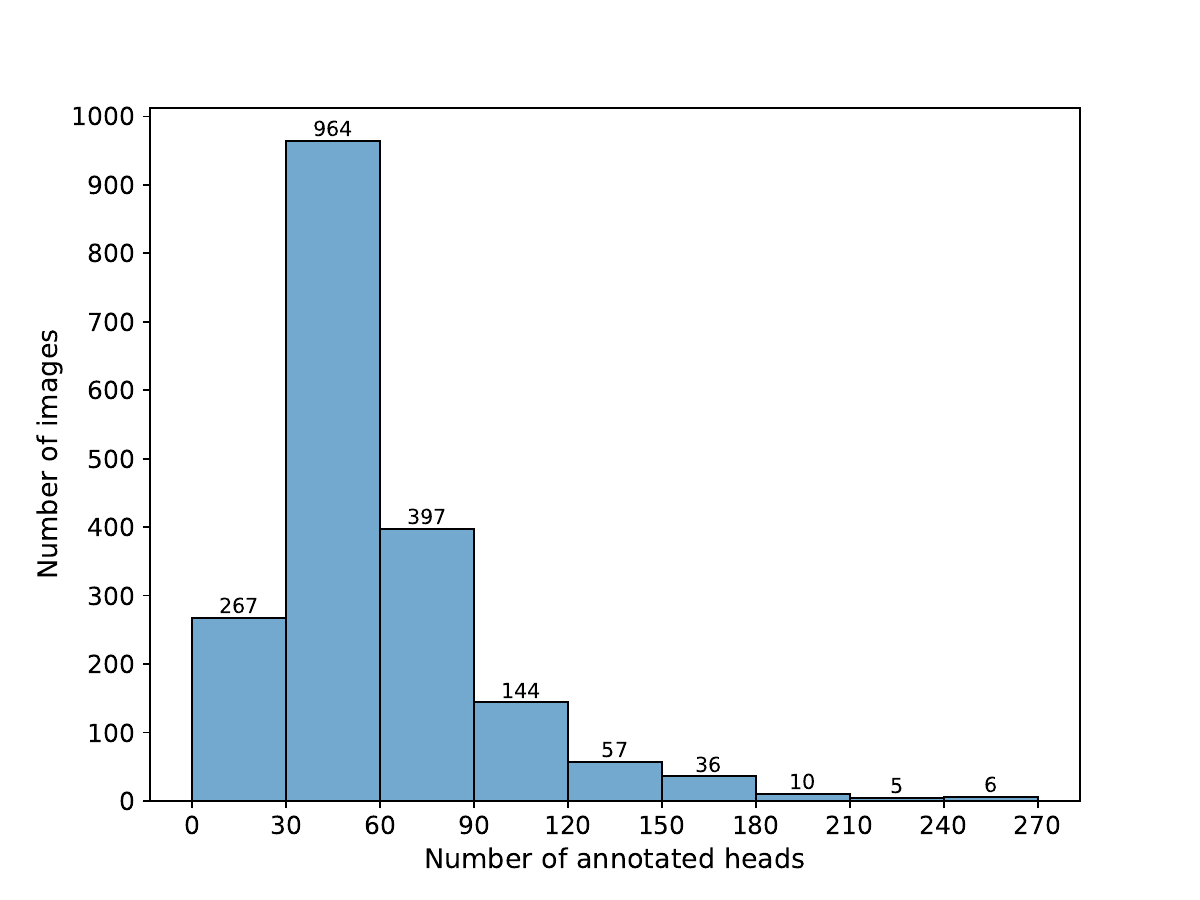}
    \caption{The distribution of the number of annotated heads among the selected images.}
    \label{fig:head_counts}
\end{figure} 

In this step, 109,913 heads were manually annotated from 1,886 frames selected from 66 video recordings.
As illustrated in~\cref{fig:head_counts}, the pedestrian count in the images within our dataset varies from 1 to 270 individuals. Specifically, 964 scenes feature between 30 and 60 persons, 397 scenes contain 60 to 90 persons, 144 frames encompass 90 to 120 persons, 57 images include 120 to 150 persons, 36 scenes range from 150 to 180 persons, and 21 scenes exhibit 180 to 270 persons. This diverse distribution enhances the dataset's comprehensiveness by encompassing various crowd densities and occlusion levels.

\begin{table}[]
\centering
\caption{Summary of training, validation, and test sets in the RPEE-Heads dataset.}
\label{tab:sets}
\resizebox{\textwidth}{!}{
\begin{tabular}{lccc|ccc|ccc}
\toprule
Set & \multicolumn{3}{c|}{Training Set} & \multicolumn{3}{c|}{Validation Set} & \multicolumn{3}{c}{Test Set} \\  
  & Frames & Heads & \% & Frames & Heads & \% & Frames & Heads & \% \\ \midrule 
Railway Platforms & 227 & 13,007 & 11.83\% & 45 & 2,492 & 2.26\% & 42 & 2,428 & 2.20\% \\  
\begin{tabular}[c]{@{}l@{}}Event Entrances (Music Concert)\end{tabular} & 715 & 41,142 & 37.43\% & 131 & 8,011 & 7.28\% & 147 & 8,333 & 7.58\% \\  
Event Entrances Experiments & 404 & 24,457 & 22.25\% & 70 & 5,519 & 5.02\% & 105 & 4,524 & 4.11\% \\ 
Overall (Total) & 1,346 & 78,606 & 71.51\% & 246 & 16,022 & 14.57\% & 294 & 15,285 & 13.90\% \\  \bottomrule
\end{tabular}
}
\end{table}

\begin{figure}[!ht]
    \centering
\includegraphics[width=0.6\linewidth]{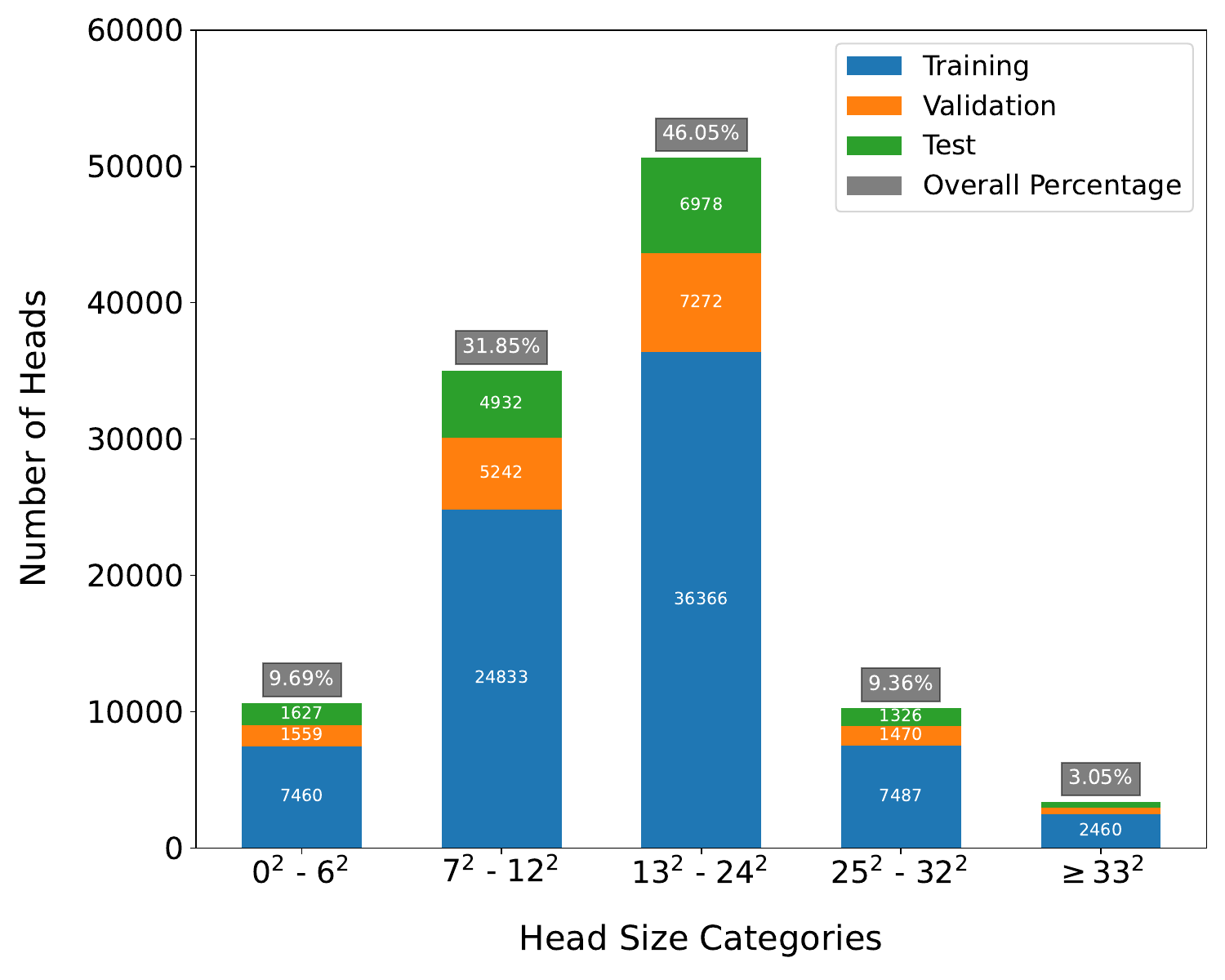}
    \caption{Distribution of the sizes of bounding boxes across the training, validation, and test sets.
 }
    \label{fig:headsizesinRPEE-Headsdataset}
\end{figure} 

 The goal of the third step is to create the annotated RPEE-Heads dataset, which includes training, validation, and test sets. To ensure diversity and minimize similarity across the sets, each video is divided into several blocks. A block consists of a segment from the sequence of selected frames ($F$). These blocks are then approximately and randomly divided into 70\% to the training set, 15\% to the validation set, and 15\% to the test set —This ratio is commonly used in DL~\cite{genc2019optimal}. Finally, the frames and their associated labeled files from these blocks make up the content of the dataset. Table~\ref{tab:sets} shows the number of frames and the annotated heads in each set of the RPEE-Heads dataset.

 The variation in head sizes or scales within the annotated dataset is a critical factor in training robust object detection algorithms, especially for identifying objects of varying dimensions. Consequently, the proposed dataset encompasses multiple size categories: less than $6^2$ pixels, between 7 and $12^2$ pixels, between $13^2$ and $24^2$ pixels, between $25^2$ and $32^2$ pixels, and greater than $32^2$ pixels. Figure~\ref{fig:headsizesinRPEE-Headsdataset} illustrates the distribution of bounding box sizes across the training, validation, and test sets. The method used to calculate these head sizes is detailed in Section 4.4. This diversity in bounding box sizes ensures the dataset's suitability for training and evaluating algorithms across different scales of head detection.
 
To summarize, the RPEE-Head dataset comprises 109,913 head annotations, with 78,606 allocated to the training set, 16,022 to the validation set, and 15,285 to the test set. The dataset was created from 1,886 frames extracted from 66 videos, recorded using various camera types and angles at railway platforms and crowded event entrances. This dataset offers diversity in weather conditions, indoor and outdoor environments, day and night times, seasons of the year, lighting conditions, head scales, crowd levels, and resolutions.

\section{Experimental 
 Evaluation}
\label{sec:experiments}
In this section, the experiments are designed with the main objective of demonstrating the effectiveness of our RPEE-Heads dataset in building robust DL models for detecting pedestrians heads at railway platforms and event entrances.

To this end, we first employ a number of well-known algorithms in this domain (namely, Fast R-CNN, Cascade R-CNN, RetinaNet-101, Faster R-CNN, RT-DETR, YOLOv7x, YOLOv8x, and YOLOv9-E) to fit model parameters using our RPEE-Heads dataset. Then, we train the top-3 performing models (YOLOv8x, YOLOv9-E, and RT-DETR) using the publicly available datasets. Finally, we conduct a comparison analysis to showcase the outstanding performance of the models built by the RPEE-Heads dataset over those models fitted based on the other datasets. 

\subsection{Implementation Details}

The generated models are trained using a supercomputer hosted in Jülich Supercomputing Center\footnote{JUWELS Booster is a multi-petaflop modular supercomputer operated by the Jülich Supercomputing Centre at Forschungszentrum Jülich. For further details, refer to https://www.fz-juelich.de/en/ias/jsc/systems/supercomputers/juwels.}. Only 4 NVIDIA A100 GPUs are utilized, each with a memory of 40 GB.

To assess the performance of the different models in terms of inference speed, a personal PC with NVIDIA RTX 3060 (12 GB GPU memory) is employed.  
Furthermore, the software packages used to train and evaluate the models are implemented using Python 3.11 along with Ultralytics~\cite{Jocher_Ultralytics_YOLO_2023}, Detectron 2~\cite{wu2019detectron2}, Official YOLOv7~\cite{wang2023yolov7}, Official YOLOv9~\cite{wang2024yolov9}, scikit-learn~\cite{scikit-learn}, PyTorch~\cite{NEURIPS2019_9015}, and OpenCV~\cite{opencv_library} libraries.
Moreover, all models are trained from scratch utilizing the Stochastic Gradient Descent optimization algorithm and their respective default hyperparameters, as detailed in~\cref{tab:hyperparameter}.

\begin{table}[ht]
\centering
 \caption{Hyperparameter settings for object detection models.}
\label{tab:hyperparameter}
\begin{tabular}{lccccc}
\toprule
Model &{Initial Learning Rate} & Final Learning Rate & Input Size & 
Iterations
& Library$^*$\\ \bottomrule
YOLOv7x & 0.01 & 0.1 & 640 & 25,000 & Official YOLOv7~\cite{wang2023yolov7} \\
YOLOv8x & 0.01 & 0.01 & 640 & 25,000 & Ultralytics~\cite{Jocher_Ultralytics_YOLO_2023}  \\
YOLOv9-E & 0.01 & 0.0001 & 640 & 30,000 & Official YOLOv9~\cite{wang2024yolov9}  \\
RT-DETR & 0.01 & 0.01 & 640 & 6,000 & Ultralytics \\
Faster R-CNN & 0.003 & 0.0003 & 1,333 & 320,000 & Detectron2~\cite{wu2019detectron2} \\
RetinaNet-101 & 0.01 & 0.0001 & 1,333 & 90,000 & Detectron2 \\
Cascade R-CNN & 0.02 & 0.002 & 1,333 & 270,000 & Detectron2 \\
Fast R-CNN & 0.001 & 0.00001 & 1,333 & 100,000 & Detectron2 \\ \hline

\multicolumn{6}{p{450pt}}{ $^*$ The library utilized for implementing, training, and evaluating the corresponding model.}\\   
\end{tabular}
\end{table}

\subsection{Evaluation Metrics}

To provide a comprehensive analysis of the effectiveness and efficiency of the trained models, several metrics were utilized: precision, recall, F1-Score (F1), mean average precision (mAP), and inference time. 

\textbf{Precision} is the ratio of correct positive detections among the instances detected as positive by the model. Formally, 
\begin{equation}
    \text{Precision} = \frac{\text{TP}}{\text{TP} + \text{FP}}
    \label{eq:precision}
\end{equation}
where TP (true positive) refers to correct positive predictions, while FP (false positive) represents incorrect positive predictions.

\textbf{Recall} is the fraction of correct positive detections among all relevant instances, as described in:
\begin{equation}
    \text{Recall} = \frac{\text{TP}}{\text{TP} + \text{FN}}
    \label{eq:recall}
\end{equation}
where FN (false negative) denotes relevant instances that the model failed to detect.

\textbf{F1-score}  is the harmonic mean of precision and recall, which provides a balanced measure of a model's performance. It is defined as:
\begin{equation}
    \text{F1}= 2 \times \frac{\text{Precision} \times \text{Recall}}{\text{Precision} + \text{Recall}}
    \label{eq:f1}
\end{equation}

The \textbf{mAP} is a popular metric used to evaluate the performance of models in object detection tasks, which is computed as in~\cref{eq:map}
\begin{equation}
  \text{mAP} =\frac{1}{n} \sum_{i=1}^{n} AP_i
  \label{eq:map}
\end{equation}
where $n$ is the number of classes, while $AP_i$ is a value that summarizes the precision-recall curve, representing the average of all precision values for each class $i$.
More specifically, the AP is calculated as the sum of precision values at each threshold, weighted by the increase in recall. Formally, AP is defined as
\begin{equation}
\text{AP} = \sum_{i=1}^{k-1} (Recall_{i+1} - Recall_{i}) \times precision_i,
\label{eq:ap}
\end{equation}
where $k$ refers to the number of thresholds.
It is essential to highlight that the Intersection over Union (IoU) threshold plays a crucial role in the above metrics, as it determines the minimum level of acceptance for the model's localization accuracy. IoU represents the degree of overlap between the predicted bounding box and the ground truth bounding box. If the prediction's IoU is greater than or equal to the IoU threshold, the detection is considered correct; otherwise, it is considered incorrect. The IoU is calculated by:

\begin{equation}
    \text{IoU} = \frac{\text{Overlap Area}}{\text{Union Area}}.
    \label{eq:IoC}
\end{equation}

All experiments in this paper utilize an IoU threshold of 0.5. This threshold is widely selected because it ensures that the predicted bounding box reasonably aligns with the ground truth, especially when dealing with small objects such as pedestrian heads~\cite{zhang2023aphid,  everingham2010pascal}.

Eventually, we evaluate the runtime of model inference by calculating the \textbf{Inference Time} which is the time (in milliseconds) that is required by the model to process an input (a frame) and produce the corresponding prediction.

\subsection{Performance of DL Algorithms on RPEE-Heads Dataset}

In this section, we compare the performance of several state-of-the-art DL models trained and evaluated using the RPEE-Heads dataset. The results of employing a number of eight DL algorithms (Fast R-CNN, Cascade R-CNN, RetinaNet-101, Faster R-CNN, RT-DETR, YOLOv7x, YOLOv8x, and YOLOv9-E) are illustrated in~\cref{tab:comparisonresults}.  Both YOLOv9-E and RT-DETR achieved the highest mAP, with about 91\% for each. Considering both the inference time and mAP, jointly we can infer that the models (RT-DETR, YOLOv7x, YOLOv8x, and YOLOv9-E) outperform others in terms of efficacy and speed, as they achieve mAP values of at least 87\% and inference time of at most 21ms.

 \begin{table}[]
    \caption{Comparative analysis of popular DL models on the RPEE-Heads dataset, sorted by mAP@0.5 (\%).}
    \centering
    \begin{tabular}{lcccccccc} \toprule
    Model &
      \multicolumn{1}{l}{Prec. (\%)} &
      \multicolumn{1}{l}{Rec. (\%)} &
      \multicolumn{1}{l}{F1 (\%)} &
      \multicolumn{1}{l}{mAP@0.5 (\%)} &
      TP &
      FP &
      FN &
      \multicolumn{1}{l}{Inf. Time (ms)} \\ \midrule
    RT-DETR       & 92   & 85  & 88 & 90.8 & 13,046 & 1,120 & 2,231 & 14  \\
    YOLOv9-E      & 91.5 & 82.9 & 87 & 90.7 & 12,672 & 1,172 & 2,606 & 11  \\
    YOLOV8x       & 88   & 81  & 84 & 88.1 & 12,488 & 1,545 & 2,789 & 17  \\
    YOLOV7x       & 89.8 & 80  & 85 & 87.4 & 12,285 & 1,388 & 2,993 & 21  \\
    Cascade R-CNN & 91   & 82  & 86 & 81.1 & 12,639 & 1,105 & 2,639 & 53  \\
    Faster R-CNN  & 90   & 80  & 85 & 81   & 12,232 & 1,355 & 3,047 & 104 \\
    Fast R-CNN    & 92   & 79  & 85 & 80.5 & 12,537 & 991  & 2,741 & 247 \\
    RetinaNet-101 & 79   & 76  & 77 & 73.8 & 11,701 & 2,947 & 3,576 & 52  \\
    \bottomrule

    \multicolumn{9}{p{420pt}}{ Prec.: precision. Rec.: recall. F1: F1-Score.  Inf.: inference. TP: true positive (correct detection). FP: false positive (incorrect detection). FN: false negative (missed detection). ms: milliseconds. }\\
    
    \end{tabular}
    \label{tab:comparisonresults}
    \end{table}
    
\begin{figure}[!ht]
    \begin{minipage}{0.48\linewidth}
        \centering
        \includegraphics[width=\linewidth]{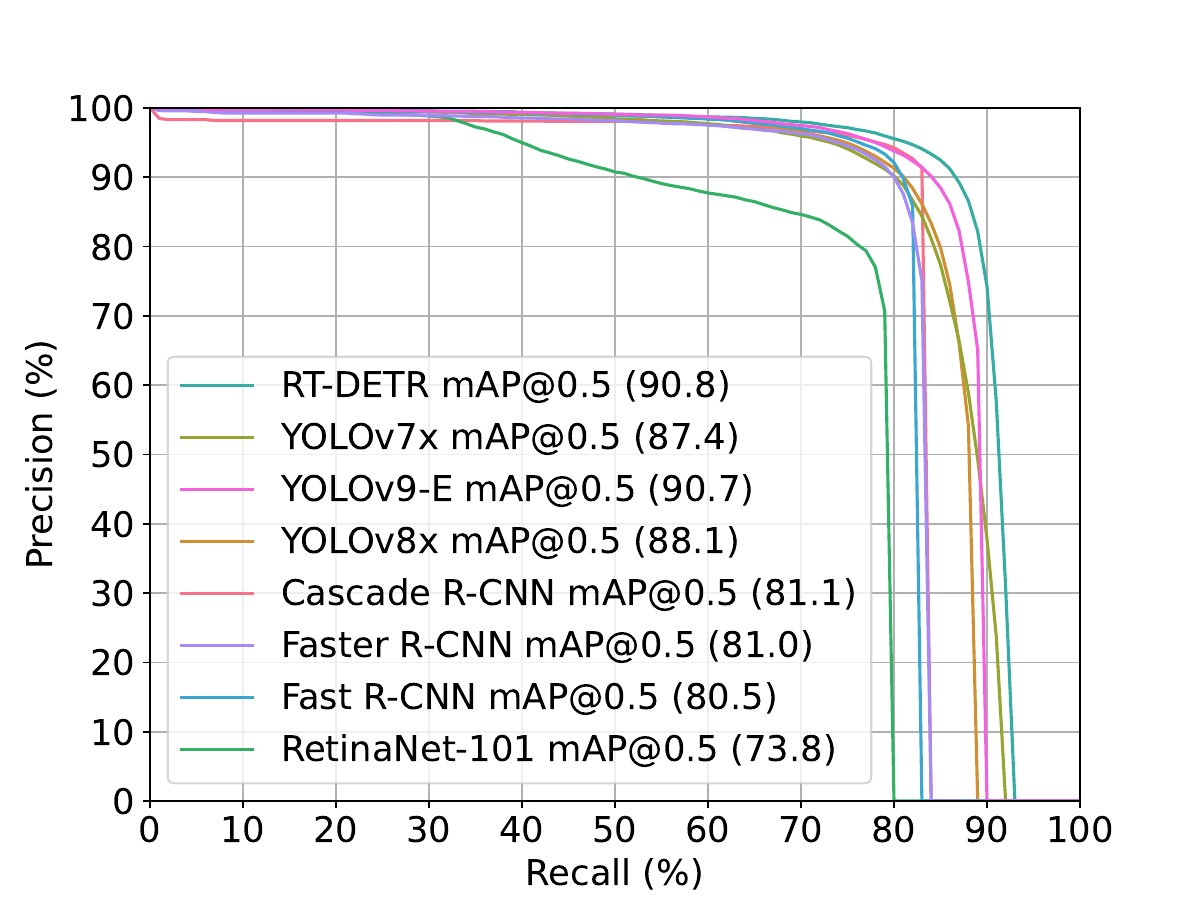}
        \caption{Performance evaluation using precision-recall curve.}
        \label{fig:mAPCombined}
    \end{minipage}\hfill
    \begin{minipage}{0.48\linewidth}
        \centering
        \includegraphics[width=\linewidth]{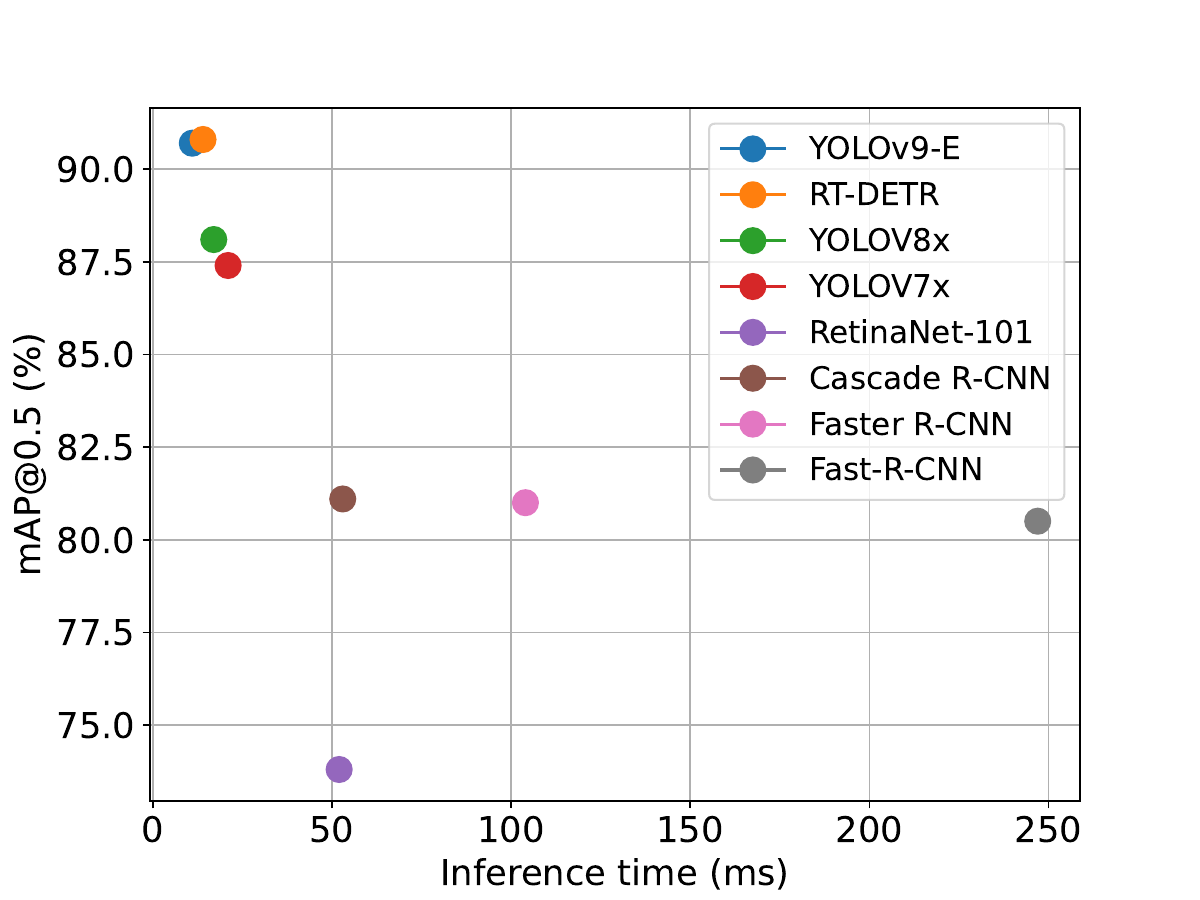}
        \caption{Mean average precision (mAP) vs. inference time.}
        \label{fig:mAPtime}
    \end{minipage}
\end{figure}

\begin{figure}[!ht]
    \centering
    \includegraphics[width=0.8\linewidth]{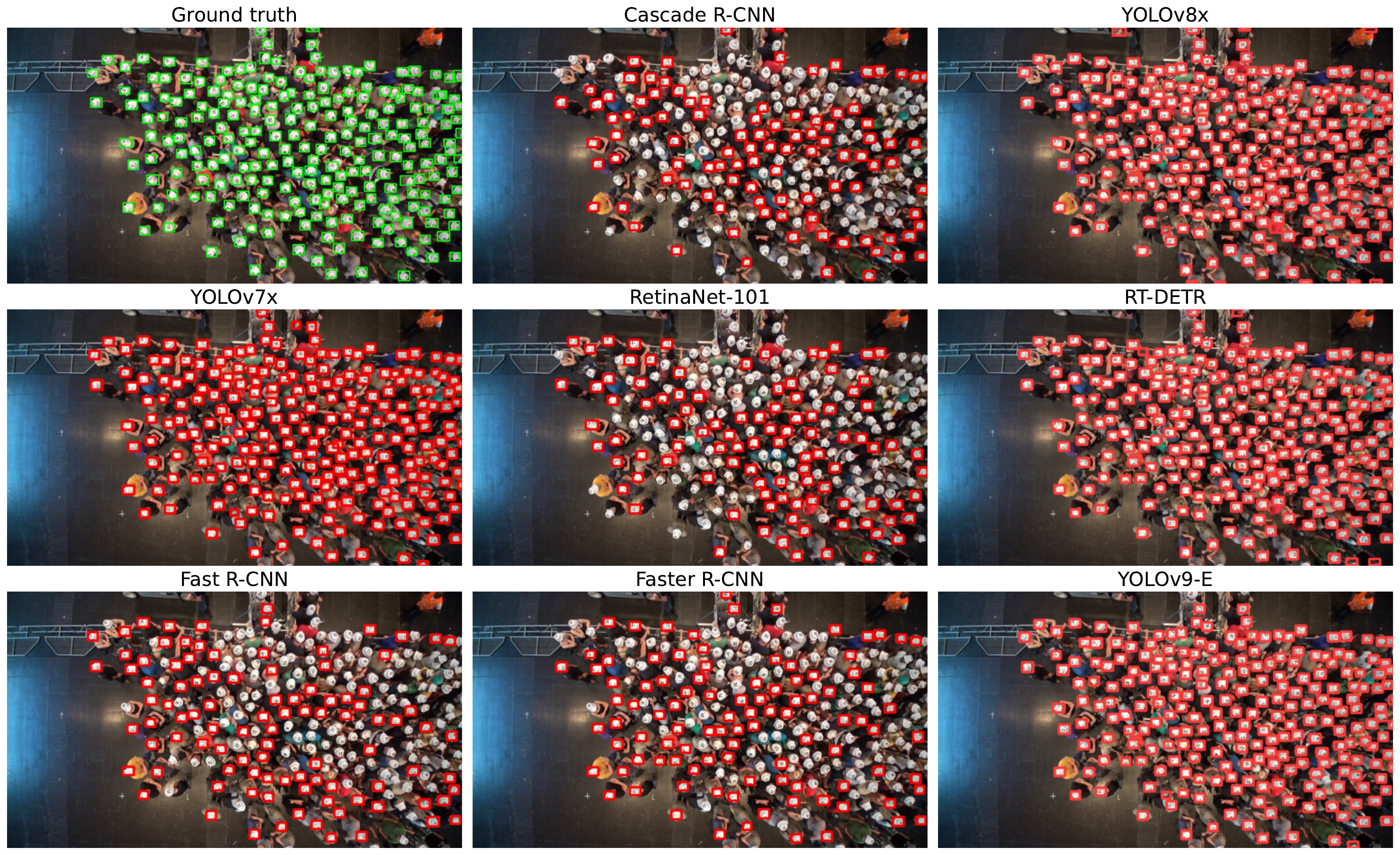}
    \caption{Visual comparison of models trained on the introduced dataset (real-world experiment scenario).}
    \label{fig:scene1}
\end{figure}

Additionally,~\cref{fig:mAPCombined} depicts the trade-offs between precision and recall at various threshold values for each  model. A model with a larger area under the precision-recall curve demonstrates better performance in distinguishing positive from negative instances, compared to models with smaller areas. To have an idea about the behavior of the different models on a single frame, a visual comparison between the ground truth and the detection results is illustrated in~\cref{fig:scene1}.
  
In summary, the DL models trained on our proposed dataset demonstrated promising performance in identifying pedestrian heads within crowds at railway platforms and event entrances, especially YOLOv9-E and RT-DETR models. However, it remains unclear whether the RPEE-Heads dataset significantly contributed to improving the accuracy of these models. Therefore, we present additional experiments to evaluate the impact of the RPEE-Heads dataset on enhancing the head detection performance in the following section.

\subsection{Influence of the RPEE-Heads Dataset and Head Size on DL Performance}
\label{sec:headsize}
This section has two objectives: 1) to study the impact of head size on the performance of DL object detection algorithms, and 2) to evaluate the effect of the proposed RPEE-Heads dataset on the performance of DL algorithms in the context of railway platforms and event entrances.

\begin{table}[H] 
\centering
\caption{Head scales for publicly available related datasets.}
\label{tab:datasets_head_scales}
\setlength{\tabcolsep}{3pt}
\begin{tabular}{ l c c c c c  c c c c c c}
\toprule
\textbf{Dataset} &  \textbf{\(0^2\text{--}6^2\) pixel$^2$ } & \textbf{\(7^2\text{--}12^2  \)pixel$^2$  } & \textbf{\(13^2\text{--}24^2  \)pixel$^2$  } & \textbf{\(25^2\text{--}32^2  \) pixel$^2$ } & \textbf{\(\geq 33^2 \)pixel$^2$  } \\
\midrule
FDST ~\cite{Fang2019}  & 4.64\% & 35.40\% & 54.30\% & 4.78\% & 0.88\% \\
SCUT-HEAD Part A ~\cite{peng2018detecting}  & 0.03\% & 21.64\% & 68.97\% & 6.95\% & 2.41\% \\
SCUT-HEAD Part B ~\cite{peng2018detecting}  & 0.03\% & 2.91\% & 23.33\% & 18.70\% & 55.03\% \\
JHU-CROWD++ ~\cite{Sindagi2022}  & 61.37\% & 19.27\% & 13.56\% & 2.81\% & 2.99\% \\
CrowdHuman ~\cite{shao2018crowdhuman}  & 20.56\% & 22.17\% & 25.80\% & 10.50\% & 20.97\% \\
CroHD ~\cite{sundararaman2021tracking}  & 0.41\% & 54.51\% & 45.06\% & 0.02\% & 0.00\% \\
NWPU-crowd ~\cite{Wang2021} & 74.40\% & 15.95\% & 6.98\% & 1.26\% & 1.41\% \\
Hollywood Heads ~\cite{vu15heads}  & 0.00\% & 0.07\% & 1.05\% & 1.28\% & 97.60\% \\
RPEE-Head  & 9.69\% & 31.85\% & 46.05\% & 9.36\% & 3.06\% \\
\bottomrule
\end{tabular}
\end{table}

To study the impact of head size on detection algorithms, we first categorize head sizes in several popular public datasets into five distinct groups: $0^{2}$–$6^{2}$ pixel$^{2}$, $7^{2}$–$12^{2}$ pixel$^{2}$, $13^{2}$–$24^{2}$ pixel$^{2}$, $25^{2}$–$32^{2}$ pixel$^{2}$, and $\geq 33^{2}$ pixel$^{2}$. Then, we  train and evaluate the performance of the top three algorithms—YOLOv9-E, YOLOv8x, and RT-DETR—over these datasets, since these algorithms are selected based on the results described in~\cref{tab:comparisonresults}. Finally, we  examine the relationship between the performance of these models and the categorized head sizes in each dataset.
Table~\ref{tab:datasets_head_scales} shows the distribution of head sizes across several datasets, while~\cref{tab:same-Test} presents the performance of the models. 
The results reveal that small head sizes, particularly those lesser than $6^{2}$ pixel$^{2}$, significantly influence the performance of DL detection algorithms. For instance, the JHU-CROWD++ dataset comprises 61.37\% of heads smaller than $6^{2}$ pixel$^{2}$, resulting in a mAP of only 22.1\% for the best model, YOLOv9-E. Similarly, the NWPU-Crowd dataset has over 74\% of heads within $6^{2}$ pixel$^{2}$, which is expected to lead to suboptimal performance.
In contrast, the CrowdHuman dataset contains 20.56\% of heads that fall within the size ranges of $0^{2}-6^{2}$ pixel$^{2}$, and it achieves a mAP of 77\%.
Other examples include the FDST dataset, which contains 4.64\% of heads smaller than $6^{2}$ pixel$^2$, SCUT-HEAD with 0.035\%, and HollywoodHead, which lacks tiny heads. In these cases, the DL models achieved over 90\% mAP. This suggests that datasets with a larger proportion of heads smaller than $6^{2}$ pixel$^2$ degrade the performance of DL detection algorithms.
These results indicate that datasets with a larger proportion of small heads, especially those smaller than $6^{2}$ pixel$^2$, degrade the performance of DL  detection algorithms. The primary reason is that small heads, particularly within cluttered and dynamic backgrounds, either lack sufficient information or make it difficult for the algorithms to extract relevant features~\cite{zhang2016far}.

\begin{table}[]
\centering

\caption{Performance of models in publicly available datasets' context. }
\label{tab:same-Test}

\begin{adjustbox}{width=0.8\textwidth}
\tiny
 \begin{tabular}{lccccc}
\toprule

{Dataset} & {Model} & {Precision (\%)} &{Recall (\%)} & {F1-Score (\%)} & {mAP@0.5 (\%)} \\
\midrule
\multirow{3}{*}{\centering FDST} & YOLOv9-E & 93.6 & 87.6 & 91.0 & 93.8 \\
 & YOLOV8x & 92.6 & 84.6 & 88.0 & 95.0 \\
 & RT-DETR & 88.0 & 89.0 & 88.0 & 96.6 \\
\midrule
\multirow{3}{*}{\centering SCUT\_HEAD\_Part\_A} & YOLOv9-E & 91.0 & 89.0 & 90.0 & 95.4 \\
 & YOLOV8x & 92.0 & 92.0 & 92.0 & 95.0 \\
 & RT-DETR & 94.2 & 93.8 & 94.0 & 96.6 \\
\midrule
\multirow{3}{*}{\centering SCUT\_HEAD\_Part\_B} & YOLOv9-E & 89.6 & 89.3 & 89.0 & 96.5 \\
 & YOLOV8x & 91.2 & 86.7 & 89.0 & 95.3 \\
 & RT-DETR & 89.9 & 89.7 & 90.0 & 96.2 \\
\midrule
\multirow{3}{*}{\centering JHU-CROWD++} & YOLOv9-E & 26.1 & 13.0 & 17.3 & 22.1 \\
 & YOLOV8x & 26.8 & 14.5 & 18.8 & 20.3 \\
 & RT-DETR & 28.4 & 10.1 & 15.0 & 16.4 \\
\midrule
\multirow{3}{*}{\centering CrowdHuman} & YOLOv9-E & 84.7 & 54.8 & 67.0 & 77.0 \\
 & YOLOV8x & 82.0 & 53.4 & 65.0 & 74.6 \\
 & RT-DETR & 79.9 & 48.7 & 61.0 & 69.0 \\
\midrule
\multirow{3}{*}{\centering HollywoodHeads} & YOLOv9-E & 93.1 & 86.6 & 90.0 & 94.0 \\
 & YOLOV8x & 93.1 & 85.5 & 89.0 & 92.9 \\
 & RT-DETR & 92.1 & 88.5 & 90.0 & 95.2 \\
\bottomrule
\end{tabular}
\end{adjustbox}
\end{table}

\begin{figure}[]
    \centering
    \includegraphics[width=1\linewidth]{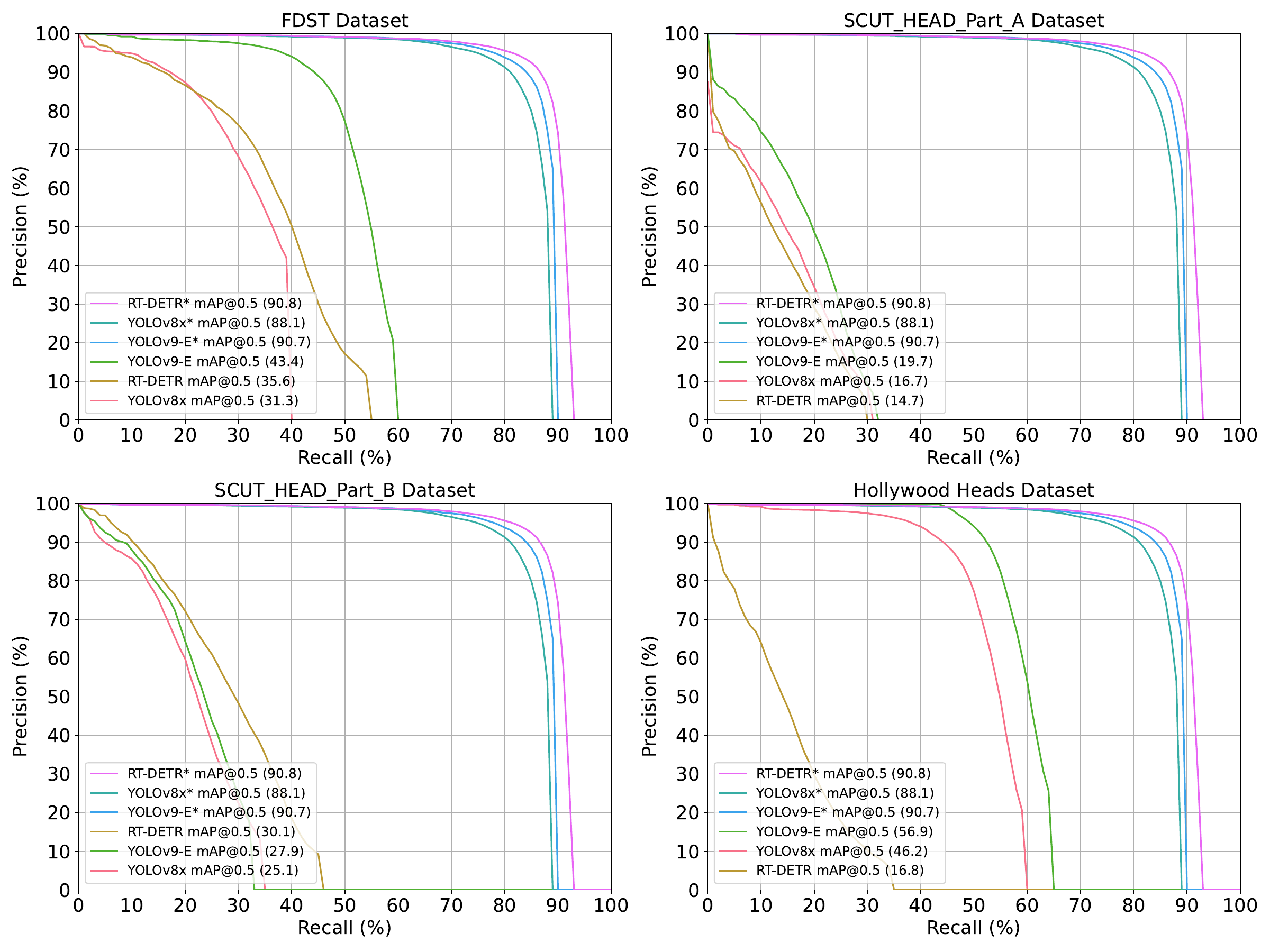}
    \caption{Precision-recall curves of diverse models and public datasets in the context of railway platforms and event entrances. * indicates models trained and evaluated using the RPEE-Heads dataset, while models without * were trained on public datasets and evaluated on the proposed dataset.}
    \label{fig:topmodelscomparisons2}
\end{figure}

To highlight the impact of our RPEE-Heads dataset on building a robust detection model for railway platforms and event entrances, we first select the FDST, SCUT-HEAD PartA, SCUT-HEAD Par B, and HollywoodHeads datasets. These datasets are efficient for head detection using DL algorithms in their contexts. Next, we train the top three DL algorithms on each selected dataset from scratch. Finally, we assess their performance using a test set from the RPEE-Heads dataset.
This is important to determine how well they perform in the context of railway platforms and event entrances.
As illustrated in~\cref{fig:topmodelscomparisons2},
these models demonstrate a significant performance gap, with a mAP value of at most 57\%, compared to the models trained on the RPEE-Heads dataset, which achieved a mAP value of at least 88\%.
This performance discrepancy can be attributed to the differences in viewpoint variations, lighting conditions, scale, occlusion, and clutter between the contexts represented by publicly available datasets and the railway platforms and event entrances~\cite{rothmeier2023had, diwan2023object}.

In summary, advanced DL-based object detection algorithms struggle to learn from objects or heads occupying less than 36 square pixels. These small objects lack discernible features, making distinguishing them from background clutter challenging. Furthermore, pedestrian head detection models trained on publicly available datasets generally perform poorly in the context of railway platforms and event entrances. In contrast, models trained using the RPEE-Heads dataset significantly outperform those generated from other datasets, demonstrating their reliability for this specific application. 

\section{Conclusion}
\label{sec:conclusion} 
This paper introduced a new pedestrian head detection benchmark within crowds at railway platforms and event entrances. Firstly, a novel, diverse, and high-resolution RPEE-Heads annotated dataset has been proposed. Two annotators manually labeled 109,913 heads in 1,886 images. Such images have been extracted from 66 videos of railway platforms and event entrances captured using various camera types and angles. The videos offer diversity in weather conditions, indoor and outdoor environments, day and night times, seasons, lighting conditions, head scales, crowd levels, and resolutions. Secondly, we trained and evaluated eight state-of-the-art DL object detection algorithms over the proposed dataset. The results showed that the mAP of the trained models for human head detection at railway platforms and event entrances ranged from 73.8\% to 90.8\%. In contrast, the best model trained on existing pedestrian head datasets achieved only 56.7\% mAP  in the context of railway platforms and event entrances. Thirdly, this paper presented an empirical study on the impact of head size on detection algorithm performance, examining five size ranges: $0^{2}$–$6^{2}$ square pixels, $7^{2}$–$12^{2}$ square pixels, $13^{2}$–$24^{2}$ square pixels, $25^{2}$–$32^{2}$ square pixels, and $\geq 33^{2}$ square pixels. The study highlights that head sizes within $6^{2}$ pixel$^2$ significantly affect detection performance in terms of mAP.

In the future, we plan to improve the generalization of pedestrian head detection by unifying public datasets and developing an efficient model that uses sequences of frames for detection.
\\
\\
\textbf{Author Contributions} \\
Initiating and designing the project: A.A.; Data sourcing and privacy: M.B.; Data curation and annotation: M.A.and Z.A.; Software: M.A. and Z.A.;  AI models training and evaluation: M.A.and Z.A.; Analysis and interpretation of results: M.A., Z.A, H.A. and A.A.; Writing—original draft preparation: M.A., Z.A., H.A. and A.A.; Writing—review and editing: M.A., Z.A., H.A., M.B. and A.A.; Supervision: A.A and H.A.; Project administration: A.A.; 
All authors have read and agreed to the published version of the~manuscript.
\\
\\
\textbf{Funding} \\
This work was funded by the German Federal Ministry of Education and Research (BMBF: funding number 01DH16027) within the Palestinian-German Science Bridge project framework,  and partially by the Deutsche Forschungsgemeinschaft (DFG, German Research Foundation) -- 491111487
\\
\\
\textbf{Institutional Review Board Statement} \\
This work involved human subjects in its research. 

For the real-world event entrance experiments,  approval of all ethical and experimental procedures and protocols was granted by the Ethics Board at the University of Wuppertal, Germany, and performed in line with the Declaration of Helsinki.

For the real-life recordings from railway platforms and music concert entrances, the data protection officer of Forschungszentrum Jülich, Germany stated that the collection and distribution of the recordings is permitted on the basis of article 6 (1) f of the European General Data Protection Regulation (GDPR).
\\
\\
\textbf{Informed Consent Statement}\\
Informed consent was obtained from all subjects involved in the real-world event entrance experiments.
\\
\\
\textbf{Data Availability}
The labeled dataset and the trained models produced in this paper are publicly available in the Pedestrian Dynamics Data Archive, hosted by Forschungszentrum Jülich. They are provided under the Attribution-ShareAlike 4.0 International license (CC BY-SA 4.0) and can be accessed at \url{https://doi.org/10.34735/ped.2024.2}.
\\
\\ 
\textbf{Acknowledgments} \\
 The authors sincerely thank Prof. Dr. Armin Seyfried for hosting and supporting the two AI internships at Forschungszentrum Jülich within the Institute for Advanced Simulation-7. We also extend our gratitude to Dr. Mohcine Chraibi for his invaluable discussions and support during the two internships associated with this paper. Special thanks to Ms. Alica Kandler for her efforts in establishing the repository for this paper in the Pedestrian Dynamics Data Archive hosted by Forschungszentrum Jülich. Finally, we deeply appreciate Helmholtz AI for providing the computing resources utilized in this work.
\\
\\
\textbf{Conflicts of Interest} \\
The authors declare no conflict of interest.

\bibliographystyle{unsrt}


\begin{thebibliography}{10}

\bibitem{brunetti2018computer}
Antonio Brunetti, Domenico Buongiorno, Gianpaolo~Francesco Trotta, and Vitoantonio Bevilacqua.
\newblock Computer vision and deep learning techniques for pedestrian detection and tracking: A survey.
\newblock {\em Neurocomputing}, 300:17--33, 2018.

\bibitem{deng2023deep}
Lijia Deng, Qinghua Zhou, Shuihua Wang, Juan~Manuel G{\'o}rriz, and Yudong Zhang.
\newblock Deep learning in crowd counting: A survey.
\newblock {\em CAAI Transactions on Intelligence Technology}, 2023.

\bibitem{lu2017trajectory}
Wei Lu, Xiang Wei, Weiwei Xing, and Weibin Liu.
\newblock Trajectory-based motion pattern analysis of crowds.
\newblock {\em Neurocomputing}, 247:213--223, 2017.

\bibitem{fan2022survey}
Zizhu Fan, Hong Zhang, Zheng Zhang, Guangming Lu, Yudong Zhang, and Yaowei Wang.
\newblock A survey of crowd counting and density estimation based on convolutional neural network.
\newblock {\em Neurocomputing}, 472:224--251, 2022.

\bibitem{alia2024novel}
Ahmed Alia, Mohammed Maree, Mohcine Chraibi, and Armin Seyfried.
\newblock A novel voronoi-based convolutional neural network framework for pushing person detection in crowd videos.
\newblock {\em Complex \& Intelligent Systems}, pages 1--27, 2024.

\bibitem{khekan2024impact}
Ahlam~R Khekan, Hadi~S Aghdasi, and Pedram Salehpour.
\newblock The impact of yolo algorithms within fall detection application: A review.
\newblock {\em IEEE Access}, 2024.

\bibitem{alia2023cloud}
Ahmed Alia, Mohammed Maree, Mohcine Chraibi, Anas Toma, and Armin Seyfried.
\newblock A cloud-based deep learning framework for early detection of pushing at crowded event entrances.
\newblock {\em IEEE access}, 11:45936--45949, 2023.

\bibitem{alia2022hybrid}
Ahmed Alia, Mohammed Maree, and Mohcine Chraibi.
\newblock A hybrid deep learning and visualization framework for pushing behavior detection in pedestrian dynamics.
\newblock {\em Sensors}, 22(11):4040, 2022.

\bibitem{alia2022exploitation}
Ahmed Alia, Mohammed Maree, and Mohcine Chraibi.
\newblock On the exploitation of gps-based data for real-time visualisation of pedestrian dynamics in open environments.
\newblock {\em Behaviour \& Information Technology}, 41(8):1709--1723, 2022.

\bibitem{ocejo2014subway}
Richard~E Ocejo and St{\'e}phane Tonnelat.
\newblock Subway diaries: How people experience and practice riding the train.
\newblock {\em Ethnography}, 15(4):493--515, 2014.

\bibitem{alia2022fast}
Ahmed Alia, Mohammed Maree, and Mohcine Chraibi.
\newblock A fast hybrid deep neural network model for pushing behavior detection in human crowds.
\newblock In {\em 2022 IEEE/ACS 19th International Conference on Computer Systems and Applications (AICCSA)}, pages 1--2. IEEE, 2022.

\bibitem{zhou2024unihead}
Hantao Zhou, Rui Yang, Yachao Zhang, Haoran Duan, Yawen Huang, Runze Hu, Xiu Li, and Yefeng Zheng.
\newblock Unihead: unifying multi-perception for detection heads.
\newblock {\em IEEE Transactions on Neural Networks and Learning Systems}, 2024.

\bibitem{lu2020semantic}
Ruiqi Lu, Huimin Ma, and Yu~Wang.
\newblock Semantic head enhanced pedestrian detection in a crowd.
\newblock {\em Neurocomputing}, 400:343--351, 2020.

\bibitem{khan2021scale}
Sultan~Daud Khan and Saleh Basalamah.
\newblock Scale and density invariant head detection deep model for crowd counting in pedestrian crowds.
\newblock {\em The Visual Computer}, 37(8):2127--2137, 2021.

\bibitem{dhillon2020convolutional}
Anamika Dhillon and Gyanendra~K Verma.
\newblock Convolutional neural network: a review of models, methodologies and applications to object detection.
\newblock {\em Progress in Artificial Intelligence}, 9(2):85--112, 2020.

\bibitem{alia2021enhanced}
Ahmed Alia and Adel Taweel.
\newblock Enhanced binary cuckoo search with frequent values and rough set theory for feature selection.
\newblock {\em IEEE access}, 9:119430--119453, 2021.

\bibitem{singh2021convolutional}
Navdeep Singh and Hiteshwari Sabrol.
\newblock Convolutional neural networks-an extensive arena of deep learning. a comprehensive study.
\newblock {\em Archives of Computational Methods in Engineering}, 28(7):4755--4780, 2021.

\bibitem{wang2023yolov7}
Chien-Yao Wang, Alexey Bochkovskiy, and Hong-Yuan~Mark Liao.
\newblock {YOLOv7}: Trainable bag-of-freebies sets new state-of-the-art for real-time object detectors.
\newblock In {\em Proceedings of the IEEE/CVF Conference on Computer Vision and Pattern Recognition (CVPR)}, 2023.

\bibitem{yolov8}
Ultralytics.
\newblock Yolov8.
\newblock \url{https://github.com/ultralytics/ultralytics}, 2023.
\newblock Available Online, Accessed on December 2023.

\bibitem{wang2024yolov9}
Chien-Yao Wang and Hong-Yuan~Mark Liao.
\newblock {YOLOv9}: Learning what you want to learn using programmable gradient information.
\newblock 2024.

\bibitem{girshick2014rich}
Ross Girshick, Jeff Donahue, Trevor Darrell, and Jitendra Malik.
\newblock Rich feature hierarchies for accurate object detection and semantic segmentation.
\newblock In {\em Proceedings of the IEEE conference on computer vision and pattern recognition}, pages 580--587, 2014.

\bibitem{girshick2015fast}
Ross Girshick.
\newblock Fast r-cnn.
\newblock In {\em Proceedings of the IEEE international conference on computer vision}, pages 1440--1448, 2015.

\bibitem{ren2015faster}
Shaoqing Ren, Kaiming He, Ross Girshick, and Jian Sun.
\newblock Faster r-cnn: Towards real-time object detection with region proposal networks.
\newblock {\em Advances in neural information processing systems}, 28, 2015.

\bibitem{cai2018cascade}
Zhaowei Cai and Nuno Vasconcelos.
\newblock Cascade r-cnn: Delving into high quality object detection.
\newblock In {\em Proceedings of the IEEE conference on computer vision and pattern recognition}, pages 6154--6162, 2018.

\bibitem{peng2018detecting}
Dezhi Peng, Zikai Sun, Zirong Chen, Zirui Cai, Lele Xie, and Lianwen Jin.
\newblock Detecting heads using feature refine net and cascaded multi-scale architecture.
\newblock {\em arXiv preprint arXiv:1803.09256}, 2018.

\bibitem{vu15heads}
Tuan{-}Hung Vu, Anton Osokin, and Ivan Laptev.
\newblock Context-aware {CNNs} for person head detection.
\newblock In {\em International Conference on Computer Vision (ICCV)}, 2015.

\bibitem{Wang2021}
Qi~Wang, Junyu Gao, Wei Lin, and Xuelong Li.
\newblock Nwpu-crowd: A large-scale benchmark for crowd counting and localization.
\newblock {\em IEEE Transactions on Pattern Analysis and Machine Intelligence}, 43, 2021.

\bibitem{Sindagi2022}
Vishwanath~A. Sindagi, Rajeev Yasarla, and Vishal~M. Patel.
\newblock Jhu-crowd++: Large-scale crowd counting dataset and a benchmark method.
\newblock {\em IEEE Transactions on Pattern Analysis and Machine Intelligence}, 44, 2022.

\bibitem{peng2024maritime}
Ling Peng, Yihong Zhang, and Shuai Ma.
\newblock Maritime small object detection algorithm in drone aerial images based on improved yolov8.
\newblock {\em IEEE Access}, 2024.

\bibitem{chen2012feature}
Ke~Chen, Chen~Change Loy, Shaogang Gong, and Tony Xiang.
\newblock Feature mining for localised crowd counting.
\newblock In {\em Bmvc}, volume~1, page~3, 2012.

\bibitem{zhang2018crowd}
Lu~Zhang, Miaojing Shi, and Qiaobo Chen.
\newblock Crowd counting via scale-adaptive convolutional neural network.
\newblock In {\em 2018 IEEE winter conference on applications of computer vision (WACV)}, pages 1113--1121. IEEE, 2018.

\bibitem{farhood2017counting}
Helia Farhood, Xiangjian He, Wenjing Jia, Michael Blumenstein, and Hanhui Li.
\newblock Counting people based on linear, weighted, and local random forests.
\newblock In {\em 2017 International Conference on Digital Image Computing: Techniques and Applications (DICTA)}, pages 1--7. IEEE, 2017.

\bibitem{lin2017focal}
Tsung-Yi Lin, Priya Goyal, Ross Girshick, Kaiming He, and Piotr Doll{\'a}r.
\newblock Focal loss for dense object detection.
\newblock In {\em Proceedings of the IEEE international conference on computer vision}, pages 2980--2988, 2017.

\bibitem{zhao2023detrs}
Yian Zhao, Wenyu Lv, Shangliang Xu, Jinman Wei, Guanzhong Wang, Qingqing Dang, Yi~Liu, and Jie Chen.
\newblock Detrs beat yolos on real-time object detection.
\newblock {\em arXiv preprint arXiv:2304.08069}, 2023.

\bibitem{vu2015context}
Tuan-Hung Vu, Anton Osokin, and Ivan Laptev.
\newblock Context-aware cnns for person head detection.
\newblock In {\em Proceedings of the IEEE International Conference on Computer Vision}, pages 2893--2901, 2015.

\bibitem{li2016localized}
Yule Li, Yong Dou, Xinwang Liu, and Teng Li.
\newblock Localized region context and object feature fusion for people head detection.
\newblock In {\em 2016 IEEE International Conference on Image Processing (ICIP)}, pages 594--598. IEEE, 2016.

\bibitem{chen2018headnet}
Gang Chen, Xufen Cai, Hu~Han, Shiguang Shan, and Xilin Chen.
\newblock Headnet: pedestrian head detection utilizing body in context.
\newblock In {\em 2018 13th IEEE International Conference on Automatic Face \& Gesture Recognition (FG 2018)}, pages 556--563. IEEE, 2018.

\bibitem{wang2017robust}
Yingying Wang, Yingjie Yin, Wenqi Wu, Siyang Sun, and Xingang Wang.
\newblock Robust person head detection based on multi-scale representation fusion of deep convolution neural network.
\newblock In {\em 2017 IEEE International Conference on Robotics and Biomimetics (ROBIO)}, pages 296--301. IEEE, 2017.

\bibitem{khan2019disam}
Sultan~Daud Khan, Habib Ullah, Mohammad Uzair, Mohib Ullah, Rehan Ullah, and Faouzi~Alaya Cheikh.
\newblock Disam: Density independent and scale aware model for crowd counting and localization.
\newblock In {\em 2019 IEEE International Conference on Image Processing (ICIP)}, pages 4474--4478. IEEE, 2019.

\bibitem{hassan2023crowd}
Maryam Hassan, Farhan Hussain, Sultan~Daud Khan, Mohib Ullah, Mudassar Yamin, and Habib Ullah.
\newblock Crowd counting using deep learning based head detection.
\newblock {\em Electronic Imaging}, 35:293--1, 2023.

\bibitem{vo2022pedestrian}
Xuan-Thuy Vo, Van-Dung Hoang, Duy-Linh Nguyen, and Kang-Hyun Jo.
\newblock Pedestrian head detection and tracking via global vision transformer.
\newblock In {\em International Workshop on Frontiers of Computer Vision}, pages 155--167. Springer, 2022.

\bibitem{Fang2019}
Yanyan Fang, Biyun Zhan, Wandi Cai, Shenghua Gao, and Bo~Hu.
\newblock Locality-constrained spatial transformer network for video crowd counting.
\newblock volume 2019-July, 2019.

\bibitem{sundararaman2021tracking}
Ramana Sundararaman, Cedric De~Almeida~Braga, Eric Marchand, and Julien Pettre.
\newblock Tracking pedestrian heads in dense crowd.
\newblock In {\em Proceedings of the IEEE/CVF conference on computer vision and pattern recognition}, pages 3865--3875, 2021.

\bibitem{shao2018crowdhuman}
Shuai Shao, Zijian Zhao, Boxun Li, Tete Xiao, Gang Yu, Xiangyu Zhang, and Jian Sun.
\newblock Crowdhuman: A benchmark for detecting human in a crowd.
\newblock {\em arXiv preprint arXiv:1805.00123}, 2018.

\bibitem{zhang2016far}
Shanshan Zhang, Rodrigo Benenson, Mohamed Omran, Jan Hosang, and Bernt Schiele.
\newblock How far are we from solving pedestrian detection?
\newblock In {\em Proceedings of the iEEE conference on computer vision and pattern recognition}, pages 1259--1267, 2016.

\bibitem{CroMA}
CroMa Project.
\newblock Crowd management in transport infrastructures (project number 13n14530 to 13n14533).
\newblock \url{https://www.croma-projekt.de/de}, 2018.

\bibitem{crowdqueue}
Data archive of experimental data from studies about pedestrian dynamics.
\newblock https://doi.org/10.34735/ped.da, 2018.

\bibitem{sieben2017collective}
Anna Sieben, Jette Schumann, and Armin Seyfried.
\newblock Collective phenomena in crowds—where pedestrian dynamics need social psychology.
\newblock {\em PLoS one}, 12(6):e0177328, 2017.

\bibitem{adrian2020crowds}
Juliane Adrian, Armin Seyfried, and Anna Sieben.
\newblock Crowds in front of bottlenecks at entrances from the perspective of physics and social psychology.
\newblock {\em Journal of the Royal Society Interface}, 17(165):20190871, 2020.

\bibitem{adrian2020influence}
Juliane Adrian, Maik Boltes, Anna Sieben, and Armin Seyfried.
\newblock Influence of corridor width and motivation on pedestrians in front of bottlenecks.
\newblock In {\em Traffic and Granular Flow 2019}, pages 3--9. Springer, 2020.

\bibitem{garcimartin2018redefining}
Angel Garcimart{\'\i}n, Diego Maza, Jos{\'e}~Mart{\'\i}n Pastor, Daniel~Ricardo Parisi, C{\'e}sar Mart{\'\i}n-G{\'o}mez, and Iker Zuriguel.
\newblock Redefining the role of obstacles in pedestrian evacuation.
\newblock {\em New Journal of Physics}, 20(12):123025, 2018.

\bibitem{Tzutalin2015}
Tzutalin.
\newblock {LabelImg}.
\newblock https://github.com/HumanSignal/labelImg, 2015.

\bibitem{genc2019optimal}
Burkay Genc and H{\"U}SEY{\.I}N Tunc.
\newblock Optimal training and test sets design for machine learning.
\newblock {\em Turkish Journal of Electrical Engineering \& Computer Sciences}, 27(2):1534--1545, 2019.

\bibitem{Jocher_Ultralytics_YOLO_2023}
Glenn Jocher, Ayush Chaurasia, and Jing Qiu.
\newblock {Ultralytics YOLO}, January 2023.

\bibitem{wu2019detectron2}
Yuxin Wu, Alexander Kirillov, Francisco Massa, Wan-Yen Lo, and Ross Girshick.
\newblock Detectron2.
\newblock \url{https://github.com/facebookresearch/detectron2}, 2019.

\bibitem{scikit-learn}
F.~Pedregosa, G.~Varoquaux, A.~Gramfort, V.~Michel, B.~Thirion, O.~Grisel, M.~Blondel, P.~Prettenhofer, R.~Weiss, V.~Dubourg, J.~Vanderplas, A.~Passos, D.~Cournapeau, M.~Brucher, M.~Perrot, and E.~Duchesnay.
\newblock Scikit-learn: Machine learning in {P}ython.
\newblock {\em Journal of Machine Learning Research}, 12:2825--2830, 2011.

\bibitem{NEURIPS2019_9015}
Adam Paszke, Sam Gross, Francisco Massa, Adam Lerer, James Bradbury, Gregory Chanan, Trevor Killeen, Zeming Lin, Natalia Gimelshein, Luca Antiga, Alban Desmaison, Andreas Kopf, Edward Yang, Zachary DeVito, Martin Raison, Alykhan Tejani, Sasank Chilamkurthy, Benoit Steiner, Lu~Fang, Junjie Bai, and Soumith Chintala.
\newblock Pytorch: An imperative style, high-performance deep learning library.
\newblock In {\em Advances in Neural Information Processing Systems 32}, pages 8024--8035. Curran Associates, Inc., 2019.

\bibitem{opencv_library}
G.~Bradski.
\newblock {The OpenCV Library}.
\newblock {\em Dr. Dobb's Journal of Software Tools}, 2000.

\bibitem{zhang2023aphid}
Tianxiao Zhang, Kaidong Li, Xiangyu Chen, Cuncong Zhong, Bo~Luo, Ivan Grijalva, Brian McCornack, Daniel Flippo, Ajay Sharda, and Guanghui Wang.
\newblock Aphid cluster recognition and detection in the wild using deep learning models.
\newblock {\em Scientific Reports}, 13(1):13410, 2023.

\bibitem{everingham2010pascal}
Mark Everingham, Luc Van~Gool, Christopher~KI Williams, John Winn, and Andrew Zisserman.
\newblock The pascal visual object classes (voc) challenge.
\newblock {\em International journal of computer vision}, 88:303--338, 2010.

\bibitem{rothmeier2023had}
Thomas Rothmeier, Diogo Wachtel, Tetmar von Dem Bussche-H{\"u}nnefeld, and Werner Huber.
\newblock I had a bad day: Challenges of object detection in bad visibility conditions.
\newblock In {\em 2023 IEEE Intelligent Vehicles Symposium (IV)}, pages 1--6. IEEE, 2023.

\bibitem{diwan2023object}
Tausif Diwan, G~Anirudh, and Jitendra~V Tembhurne.
\newblock Object detection using yolo: Challenges, architectural successors, datasets and applications.
\newblock {\em multimedia Tools and Applications}, 82(6):9243--9275, 2023.

\end{thebibliography}
\end{document}